\pdfoutput=1

\documentclass[11pt]{article}

\usepackage{ACL2023}

\usepackage{times}
\usepackage{latexsym}

\usepackage[T1]{fontenc}

\usepackage[utf8]{inputenc}

\usepackage{microtype}

\usepackage{inconsolata}

\usepackage{graphicx}

\title{Case-Based Reasoning Approach for Solving Financial Question Answering\\
(March.29, 2024)}

\author{Yikyung Kim \\
  Seoul National University \\
  \texttt{k2y1513@snu.ac.kr} \\\And
  Jay-Yoon Lee \\
  Seoul National University \\
  \texttt{lee.jayyoon@snu.ac.kr} \\}

\begin{document}
\maketitle

\begin{abstract}
Measuring a machine’s understanding of human language often involves assessing its reasoning skills, i.e. logical process of deriving answers to questions. While recent language models have shown remarkable proficiency in text-based tasks, their efficacy in complex reasoning problems involving heterogeneous information such as text, tables, and numbers remain uncertain. Addressing this gap, FinQA (\citep{chen2021finqa}) introduced a numerical reasoning dataset for financial documents and simultaneously proposed a program-generation approach . Our investigation reveals that half of the errors (48\%) stem from incorrect operations being generated. To address this issue, we propose a novel approach to tackle numerical reasoning problems using case-based reasoning (CBR), an artificial intelligence paradigm that provides problem-solving guidance by offering similar cases (i.e. similar questions and corresponding logical programs). Our model retrieves relevant cases to address a given question, and then generates an answer based on the retrieved cases and contextual information. Through experiments on the FinQA dataset, we demonstrate competitive performance of our approach and additionally show that by expanding case repository, we can help solving complex multi-step programs which FinQA showed weakness of. 
\end{abstract}

\section{Introduction}
Reading comprehension is a key metric for assessing an individual's ability to interpret and integrate written text. Natural Language Processing (NLP), developed to process large volumes of sequential data, includes the area of Question Answering (QA), which focuses on systems that automatically answer natural language questions, effectively measuring a machine's language understanding. QA research has evolved significantly since the 1960s~\citep{androutsopoulos1995natural}, particularly with deep learning's rise, marked by the creation of over 80 new benchmark datasets in the last two years~\citep{rogers2023qa}.

The advancement of large language models has notably enhanced text-based QA tasks, even outperforming humans on some benchmarks~\citep{liu2019roberta, lan2019albert, clark2020electra}. This success has spurred interest in table-based QA, leading to new frameworks for tabular data pre-training. However, despite the achievements in text-based applications, the performance in table-based QA tasks has not been as impressive~\citep{herzig2020tapas, yin2020tabert}, raising questions about their effectiveness in handling complex reasoning in multi-modal documents that include text, tables, and numerical data. The current research tends to overlook the complexities of processing different types of information found in real-world documents.

To bridge this gap, recent efforts have focused on creating new QA datasets that incorporate both tabular and textual data, necessitating numerical reasoning~\citep{chen2021finqa, zhao2022multihiertt, zhu2021tat}. Specifically, the FinQA dataset by Chen et al. (2021) provides a benchmark for numerical reasoning in financial documents~\citep{chen2021finqa}. Despite utilizing pre-trained language models such as BERT and RoBERTa, fine-tuned for these datasets, baseline models have yielded underwhelming performance. This discrepancy in results, especially given these models' effectiveness in traditional text-based QA tasks, suggests a pressing need for further research into numerical reasoning QA that integrates both table and textual data. \newline
Moreover, when evaluating the capabilities of large language models (LLM) like GPT-3.5, it becomes apparent that LLMs also face challenges in this domain. They frequently struggle to retrieve table cells and fail to comprehensively address queries requiring complex reasoning steps, underscoring the necessity for advancements in our approach to QA tasks involving numerical reasoning.

Our analysis indicates that a significant source of errors in current baseline model arises from incorrect operations during the reasoning steps. Specifically, the accuracy of producing the correct answer (execution accuracy) was often lower than the accuracy of generating the correct program (program accuracy), due to the model generating incorrect operations but accidentally arriving at the right answer. To accurately assess a machine's reasoning ability, program accuracy is deemed more crucial than execution accuracy. The current model utilizes a sequence-to-sequence (seq-to-seq) architecture with an LSTM decoder, which creates programs in an autoregressive manner, leading us to wonder if pre-informing the model about operations could be beneficial.

In response to these challenges, we propose a novel strategy that employs case-based reasoning (CBR), an artificial intelligence technique that solves new problems by referencing solutions to similar problems. Our approach involves retrieving cases that are closely related in terms of questions and programs to address specific queries, followed by producing answers leveraging these cases and contextual clues. By integrating similar questions and their operations as additional data, we aim to enhance the model's ability to solve complex reasoning tasks. Furthermore, we anticipate that providing the model with a more closely related cases will enhance its ability to generate accurate programs.

We plan to evaluate our approach across multiple settings: the basic (FinQA baseline model), gold case, and retrieved case. In the gold case environment, we provide cases with identical operations to the answer for the program generation model. After assessing the effectiveness of our strategy with gold cases, we proceed to test our CBR method using cases that have been actively retrieved. Our experiments will explore various retrieval models and methods of case presentation to the program generation model, aiming to demonstrate that our model can identify relevant cases and outperform the current baseline model.

\section{Preliminaries}
\label{sec:preliminaries}
\subsection{Overview of FinQANet}
FinQANet is designed as a benchmark model for the FinQA challenge, built around a dual-component architecture: a context retrieval system and a program generation system. Financial reports, often exceeding 2,000 tokens, pose a significant challenge for existing QA models due to their length. To address this, FinQANet begins by pinpointing the most pertinent sections of the input document. For handling tabular data, it converts each row into a readable sentence format. For instance, a table cell with row header named 'risk-free interest rate' and column header named '2006' can be converted into “the risk-free interest rate of 2006 was 5\%.” 

For a question \emph{Q} and a document \emph{D} that includes both text and tables, the model identifies the top \emph{k} relevant contexts $\emph{E}$. It then formulates a series of programmatic steps $\emph{P}=\left\{\emph{$p_0$}, \emph{$p_1$},..., \emph{$p_n$}\right\}$, where each \emph{$p_i$} represents a predefined operation or one derived from the contexts. This series of steps, when executed, produces the answer \emph{A}, calculated as follows:
\begin{equation}
P(A|Q,D)=\sum_{i}P(P_i|Q,E)
\end{equation}

\paragraph{Context Retriever}
Given the document's extensive text and tables, a context retriever was initially developed to identify the most relevant contexts for answering a specific question. The goal of this model is to pinpoint the top-\emph{k} relevant contexts \emph{E}, given a question \emph{Q} and a document \emph{D}. FinQANet utilizes a dense vector spaces approach for quickly and accurately extracting pertinent contexts from large documents, shifting from traditional, less efficient vector space models like TF-IDF and BM25~\citep{sparck1972statistical, robertson1995okapi}. This approach facilitates more meaningful semantic comparisons, leading to more precise matches between questions and their relevant contexts~\citep{karpukhin2020dense}. \newline
The context retriever leverages a pre-trained BERT model, processing combinations of the question with each context as input to the model. It then fine-tunes this classifier on the FinQA dataset. The top-\emph{k} relevant contexts identified are subsequently used in the program generation phase.

\begin{figure*}
    \centering
    \includegraphics[width=\linewidth, trim={3cm 5cm 4cm 5cm},clip]{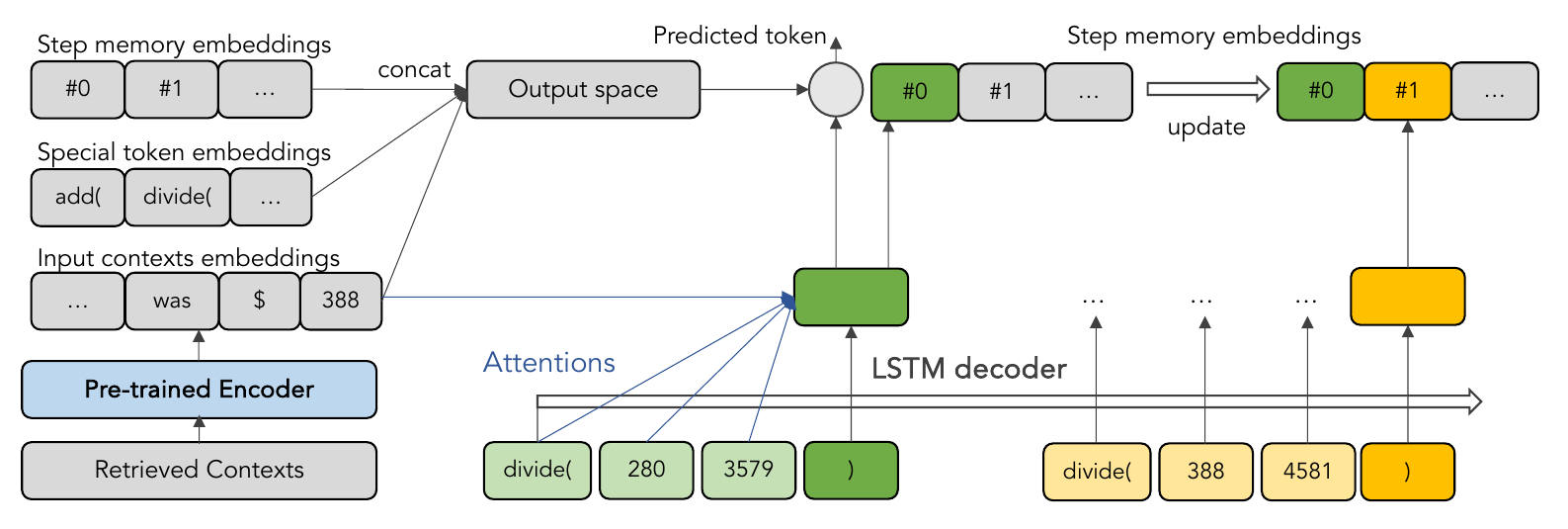}
    \caption{FinQANet Program Generator}
    \label{fig:finqa_program_generator}
\end{figure*}

\paragraph{Program Generator}
The program generator operates on a sequence-to-sequence (seq-to-seq) architecture, which is ideally suited for handling long sequential data through an encoder and decoder structure~\citep{sutskever2014sequence}. Initially, this system examines the given question \emph{Q} and the contexts \emph{E} retrieved earlier, with its encoder capturing essential information. Subsequently, the decoder systematically generates the final program \emph{P} in an autoregressive manner. For the decoding process, FinQANet employs an LSTM model, incorporating an attention mechanism to improve the precision and clarity of its outputs~\citep{bahdanau2014neural}.\newline
A key distinction of this program generator from standard seq-to-seq models lies in wider range of decoder outputs. This not only includes the input embeddings but also specific tokens for developing a logical program, such as various operations and step memory tokens. These step memory tokens are instrumental in outlining the program's logical sequence. An illustrative Figure ~\ref{fig:finqa_program_generator} details the process of program decoding within the FinQANet’s program generator.

\paragraph{Program}
The FinQA dataset utilizes a specific programming language designed for executing mathematical and tabular operations. The program sequence includes a variety of operations, such as addition, subtraction, multiplication, division, and others, tailored for table summarization and averaging. The operands for these operations are either directly extracted from the document or derived from preceding steps. For example, to perform a subtraction on the results of two division operations, it could be articulated as follows: Divide(arg1, arg2), Divide(arg1, arg2), Subtract(\#0, \#1).

\subsection{Exploring Case-Based Reasoning (CBR)}
Our research adopts a case-based reasoning (CBR) methodology, a problem-solving approach that involves a three-step cycle: (i) identifying and retrieving relevant previous cases similar to the current problem, (ii) reusing these cases to solve the problem at hand, and (iii) revising the solution if necessary for better alignment with the problem requirements~\citep{aamodt1994case, das2021case}. Our implementation focuses on the first two steps: retrieval and reuse, which will be elaborated upon in methodology.

\paragraph{Defining Gold Cases}
The quantity of potential cases for each query is equivalent to the dataset's size. Considering the extensive pool of candidates, it's essential to identify "gold cases" for model training purposes. During the training phase, we leverage those gold cases, which are ideal examples that our case retriever aims to identify. Developing a mechanism for accurately retrieving these cases poses a significant challenge. To define and select gold cases, we first evaluate the similarity between the given question and candidate question.\\
We utilize a BERT encoder to generate representations of the questions~\citep{devlin2018bert}. Through computing the cosine similarity between these vector representations, we can identify cases that closely match the input question in terms of their content and context. Cosine similarity measures the cosine of the angle between two vectors, indicating how closely related they are in terms of orientation in a multi-dimensional space. It is defined as:
\begin{equation}
Cosine Similarity = \frac{A \cdot B}{||A||  ||B||}
\end{equation}

Following the question similarity assessment, we rank candidate cases based on the similarity of their logical solutions or "programs," using program scores. To determine program similarity, we employ the Levenshtein (edit) distance, which quantifies the minimum number of edits (insertions, deletions, substitutions) needed to transform one sequence into another~\citep{levenshtein1966binary}. This comparison is done at the word level, prioritizing the matching of operations over arguments due to their critical role in logical reasoning. The program score is calculated as follows:
\begin{equation}
S = \frac{l_{ops}-d_{ops}}{l_{ops}} \cdot w_{ops}+\frac{l_{arg}-d_{arg}}{l_{arg}} \cdot w_{arg}
\end{equation}
where $l_{ops}$ and $l_{arg}$ indicate the number of operations and arguments in the target program, respectively, and $d_{ops}$ and $d_{arg}$ denote the edit distances for operations and arguments between the candidate and target programs.
Candidates with program scores above 0.9 are considered gold cases, while others form the negative set. From analyzing the top 100 candidates based on question similarity, approximately 5.3\% of our dataset did not have appropriate positive candidates. However, we expect that our case retrieval model can still effectively learn from instances that do have positive matches. We plan to explore diverse strategies to introduce more variability into our candidate pool for training purposes, which will be elaborated upon in the methodology section. The selection of the 0.9 threshold was aimed at ensuring the high quality of candidate selection.

\section{Related Works}
\label{sec:related_works}
The research area of Question Answering (QA) encompasses a wide range of tasks, including text-based, table-based, and numerical reasoning QA. Each of these subdomains contributes unique challenges and insights, laying the groundwork for our approach to addressing financial QA tasks through Case-Based Reasoning (CBR).

\subsection{Text-Based Question Answering}
Text-based QA systems have seen significant advancements, largely due to the development of large-scale datasets such as SQuAD (Stanford Question Answering Dataset) and deep learning models like BERT and its variants. These systems primarily focus on extracting answers from textual data, where the answer to a question is a segment of text found within a provided document. Rajpurkar et al. introduced SQuAD, which catalyzed progress in this area by challenging models to predict answer spans within passages~\citep{rajpurkar2016squad}. \newline
Following this, Devlin et al. presented BERT, showcasing its ability to understand the context of words in text by pre-training on a large corpus and fine-tuning on specific tasks such as SQuAD~\citep{devlin2018bert}. This methodology has set a precedent for text-based QA, influencing subsequent research and development in NLP.

\begin{figure*}[ht]
    \centering
    \includegraphics[width=\linewidth, trim={0cm 4.5cm 2cm 5.7cm},clip]{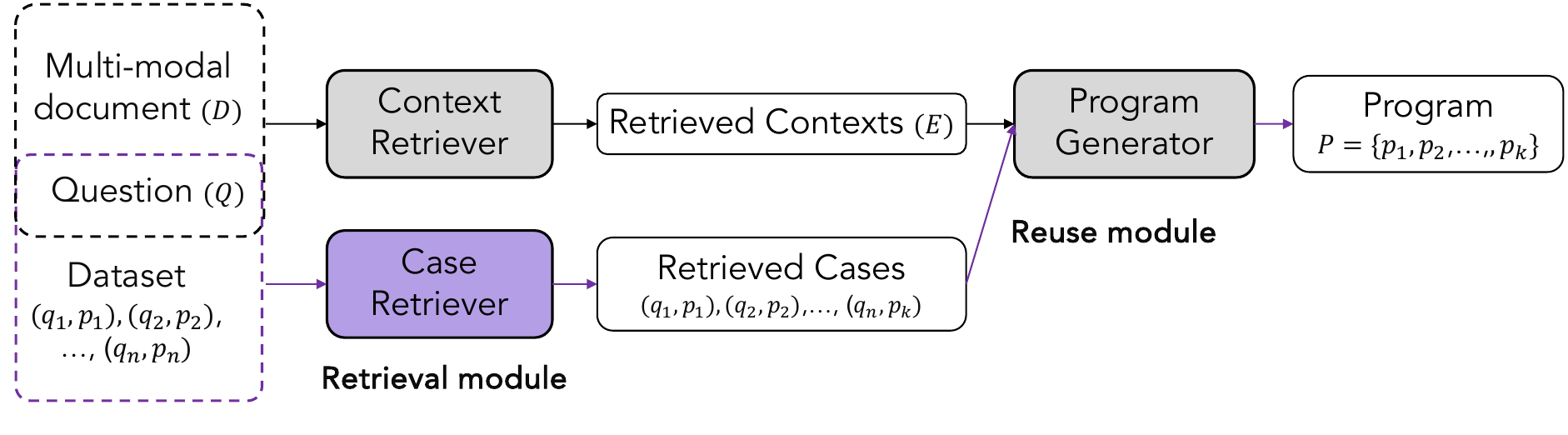}
    \caption{Overview of the Enhanced CBR Architecture. This diagram illustrates the integration of a case retriever into the FinQA baseline model. The case retriever is designed to identify relevant cases from the dataset that share similarities in question and program. These cases, along with the related contexts retrieved from the document, are then fed into the program generator to facilitate the generation of program sequence.}
    \label{fig:overall_architecture}
\end{figure*}

\subsection{Table-Based Question Answering}
Table-based QA represents a challenging segment of NLP research that demands an intricate understanding of both language and the structured nature of tables. Unlike textual data, tables organize information in a two-dimensional structure, using rows and columns to systematically categorize data. This organization facilitates direct comparison and lookup of numerical values, making tables a unique form of data representation that models must effectively interpret~\citep{dong2022table}. \newline
The WikiTableQuestions dataset, introduced by Pasupat and Liang, serves as a pivotal benchmark for evaluating the performance of table-based QA systems~\citep{pasupat2015compositional}. This dataset highlights the intricate nature of table-based queries, which often require a nuanced understanding of both a table's layout and its contained information.\newline
In response to the complexities associated with table-based QA, there has been significant progress in the development of transformer-based language models tailored to understand the dual nature of tables as both textual and structural entities. These models are designed to comprehend tables by jointly learning their semantic content and structural layout, building upon the foundational text understanding capabilities of traditional NLP models. Among these, TaPas, developed by Herzig et al., stands out as a BERT adaptation pre-trained specifically for table comprehension. TaPas enhances BERT's capabilities by introducing specialized embeddings to capture row and column information and incorporating additional classification layers for more precise cell and operation selection tasks~\citep{herzig2020tapas}. Another notable contribution is TaBERT by Yin et al., which approaches table encoding by treating tables as sequences analogous to text. This methodology enables TaBERT to produce context-aware cell representations, effectively capturing the relationships between cells and their collective relevance to a query~\citep{yin2020tabert}. Together, these developments underscore the growing importance of dedicated models and datasets in pushing the boundaries of table-based QA research.

\subsection{Question Answering Requiring Numerical Reasoning}
Numerical reasoning in QA involves questions that require arithmetic operations or other forms of mathematical reasoning, often combining textual and tabular data. This area is particularly relevant to financial QA tasks, where answers depend on calculations based on data extracted from documents. Chen et al. developed FinQA, a dataset and benchmark specifically for financial question answering that necessitates numerical reasoning over financial reports~\citep{chen2021finqa}. \newline
FinQA is built around the financial reports of S\&P 500 companies, with its questions formulated by experts in the financial domain. This setup presents a significant challenge for current models, testing their capacity to decode domain-specific knowledge and execute complex numerical reasoning. \newline
In their exploration, Chen et al. assessed various architectures to identify what is essential for effectively addressing the FinQA dataset. For instance, their examination included a combination of retrieval mechanisms and direct answer generation models. However, this approach yielded almost no success in terms of execution accuracy, underscoring the critical need for generating intermediary reasoning steps or programs. They also evaluated the performance of a combination of retrieval systems with the Neural Symbolic Reader (NeRd), which had previously achieved leading results on the MathQA dataset~\citep{chen2019neural, amini2019mathqa}. Unlike FinQANet, NeRd attempts to learn the structure of reasoning programs in a nested sequence format without the use of step memory tokens like \#0, \#1. The comparison revealed that while NeRd's performance was commendable, it still fell short of the benchmarks set by FinQANet, suggesting that understanding and learning the logical program structure offers considerable advantages.\newline
This exploration into different methodologies and their performance on the FinQA dataset clearly outlines the existing gaps in the ability of current models to handle intricate calculations and logical reasoning. It provides a strong foundation for our research, where we propose leveraging a case-based reasoning (CBR) approach to enhance the capabilities of financial QA systems.

\section{Methodology}
\label{sec:methodology}
This paper is dedicated to addressing the complexities of financial question answering (QA) through a case-based reasoning (CBR) framework. Central to our methodology are two integral components: a retrieval module for identifying similar cases and a reuse module for applying these cases to solve the problem at hand. The design of our methodological approach is visually represented in Figure \ref{fig:overall_architecture}, highlighting the process from case retrieval based on specific queries to their application in aiding the program generator. We have implemented distinct methodologies for each segment, as detailed in the subsequent sections.

\subsection{Retrieving Cases}
The number of candidate cases for each question corresponds to the size of the dataset. Given the vast number of potential cases, our objective is to filter out only the most relevant cases that will significantly aid the program generation process. To achieve this, we have devised a retrieval model capable of selecting cases with a high degree of relevance to the posed question. Throughout the training phase, we target 'gold cases'—cases that are predefined as highly relevant based on their operational similarity to the query, as mentioned in the preliminary section, with a program score above 0.9 serving as the benchmark for relevance. These gold cases form our positive set, with the remainder being classified as negative. To discriminate between these sets, we employ a classifier trained for this specific purpose, utilizing either Bi-encoders or Cross-encoders for sequence comparison, which are detailed further below.

\subsubsection{Bi-Encoder Architecture}
\begin{figure}[ht]
    \includegraphics[width=\linewidth, trim={7cm 6.5cm 7cm 6cm}, clip]{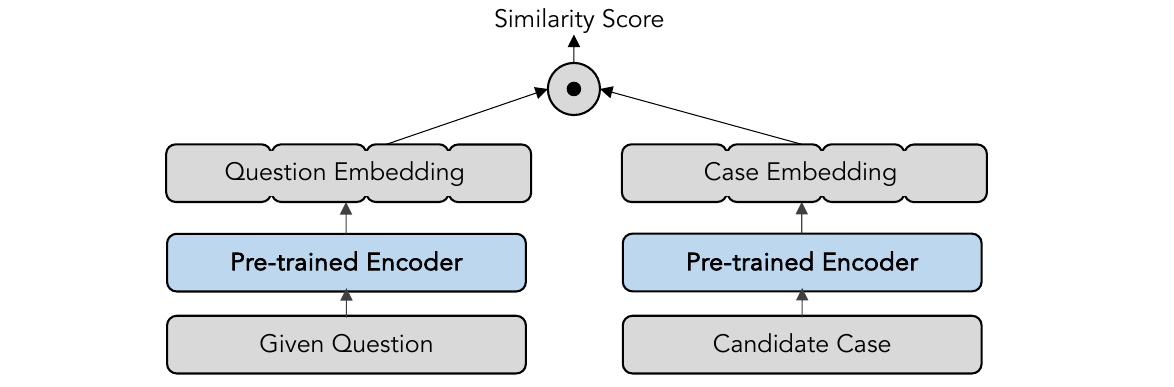} 
    \caption{Bi-Encoder Architecture: the question and candidate case are encoded separately to generate distinct embeddings.}
    \label{fig:biencoder}
\end{figure}

The bi-encoder architecture operates by independently encoding the query and potential cases into dense vector representations, as depicted in Figure \ref{fig:biencoder}. This approach utilizes separate instances of pre-trained models such as BERT or RoBERTa for encoding, ensuring that each text input, whether a query or a case, is transformed into a high-dimensional space where semantic similarities can be quantitatively assessed. Renowned for its efficiency, this approach has demonstrated commendable performance in QA tasks~\citep{mazare2018training, dinan2018wizard, karpukhin2020dense}. \newline
The strength of the bi-encoder lies in its efficiency; by pre-computing and caching the embeddings of the potential cases, the system can rapidly compare these embeddings with those of incoming queries. This characteristic makes it especially useful for rapid retrieval in extensive datasets. However, it's important to note that while bi-encoders excel in speed, they may not capture the nuanced interplay between queries and cases as effectively as cross-encoders due to their separate encoding processes~\citep{humeau2019poly}.

\subsubsection{Cross-Encoder Architecture}
In contrast to the bi-encoder, the cross-encoder architecture simultaneously encodes the query and each potential case by concatenating them into a single input sequence for processing by models like BERT or RoBERTa. The integrated encoding process allows the model to perform self-attention over the entire input, enabling it to capture subtle semantic relationships and contextual nuances between the query and the case~\citep{wolf2019transfertransfo, vig2019comparison}. \newline
While cross-encoders demonstrate superior accuracy and a deeper understanding of the query-case relationship due to their comprehensive encoding strategy, they are computationally more intensive~\citep{humeau2019poly}. Encoding each query-case pair individually results in slower retrieval times compared to bi-encoders. Nonetheless, the enhanced accuracy and contextual sensitivity of cross-encoders make them invaluable for tasks where precision is paramount, and the complexity of the financial domain often necessitates such an approach.

\begin{figure}
    \centering
    \includegraphics[scale=0.5, trim={12cm 7cm 12cm 7cm}, clip]{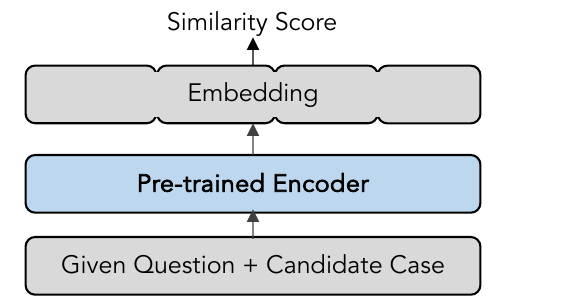}
    \caption{Cross-Encoder Architecture: the question and candidate case are concatenated before encoding to create a single embedding.}
    \label{fig:crossencoder}
\end{figure}

\paragraph{Implementation}
In our case retrieval component, we explore both architectures to balance their benefits and drawbacks. After evaluating both methods, we will select or combine them based on their performance and suitability for our model. The initial broad filtering capability of the bi-encoder, followed by the detailed scrutiny provided by the cross-encoder, could potentially offer a synergistic approach, leveraging the strengths of both models. This dual-stage process is anticipated to optimize case retrieval for enhancing the program generation phase, especially during testing, by preliminarily narrowing down relevant cases with the bi-encoder and then applying the cross-encoder for finer selection.

\subsection{Reusing Retrieved Cases in Program Generation}
After successfully retrieving relevant cases through our case retrieval module, the next pivotal step in our methodology involves effectively incorporating these cases into the program generation process. To achieve this, we draw inspiration from the program generator model used in the FinQA baseline but introduce novel adaptations by employing two distinct methods to integrate the retrieved cases: the Concatenation approach and the Separate Encoder approach. Both methods are grounded in well-established practices within the field of NLP and have shown promise in enhancing model performance by providing additional relevant context.

\subsubsection{Concatenation Approach}
The Concatenation approach is straightforward yet effective; it involves appending the retrieved cases directly to the retrieved contexts using a special separation token. This aggregated input is then processed through a pre-trained encoder, such as BERT or RoBERTa, to produce a comprehensive embedding that captures both the contextual nuances of the query and the supportive information from the cases. Following the encoding, an LSTM model takes over to sequentially generate the program tokens. This method leverages the powerful contextual embedding capabilities of pre-trained models to enhance the relevance and accuracy of the generated programs by ensuring that the additional case information is seamlessly integrated into the input sequence. The simplicity of this approach, combined with the proven effectiveness of encoder models like BERT and RoBERTa in handling concatenated inputs~\citep{wolf2019transfertransfo}, makes it a viable strategy for enriching the program generation process.

\subsubsection{Separate Encoder Approach}
In contrast, the Separate Encoder approach utilizes an additional pre-trained encoder, such as BERT or RoBERTa, dedicated exclusively to encoding the retrieved cases. This method allows for a more isolated analysis of the cases, ensuring that their content is thoroughly understood and represented before being considered in the program generation phase. The LSTM model, responsible for producing the program tokens, incorporates a separate attention mechanism directed towards the encoder output for the cases. This targeted attention enables the program generator to dynamically weigh the relevance of the case information at each step of the token generation process, potentially leading to more nuanced and accurate program outputs. By maintaining a distinct processing stream for the cases, this approach aims to maximize the utility of the retrieved information, ensuring that each case contributes meaningfully to the final program generated. \newline
Employing these approaches within the context of financial QA tasks, especially when integrated with a CBR framework, signifies a novel adjustment designed to capitalize on insights from specific cases to refine the process of generating programs. By exploring different techniques for case retrieval and program generation, our objective is to identify and apply the optimal approach for integrating detailed case information into the program creation process. This will, in turn, enhance the model's capability in addressing complex financial questions.

\section{Experiment}
\label{sec:experiment}
This section outlines the experimental framework designed to evaluate the effectiveness of our case-based reasoning (CBR) system in the area of financial question answering (QA). The primary goal of this paper is to enhance the quality of program generation in financial QA tasks. As depicted in Table \ref{initial_results_generation}, our experiments demonstrate a significant improvement in the program generator's performance when it is fed with gold cases. We hypothesize that the program generator will produce better outcomes as the case retriever's efficiency improves and supplies higher quality cases to the generator.

\begin{table*}
  \centering
  \begin{tabular}{l|ccc}
    \hline
    \textbf{Model Architecture} & \textbf{Exe Acc} & \textbf{Prog Acc} & \textbf{Ops Acc}\\
    \hline
    FinQANet (RoBERTa-base)                & 56.10         & 54.38            & -   \\
    CBR-Gold (RoBERTa-base, concat)             & 62.51         & 61.11            & 93.72  \\
    CBR-Gold (RoBERTa-base, separate-encoder)   & 59.02         & 57.45            & 86.39 \\
    CBR-Retrieved-q (RoBERTa-base, concat)      & 54.14         & 51.78            & 75.67 \\
    CBR-Retrieved-qp (RoBERTa-base, concat)     & 58.15         & 55.71            & 77.68 \\
  \hline
  \end{tabular}
  \caption{Initial Experimental Results on Program Generation}
  \label{initial_results_generation}
\end{table*}

\subsection{Dataset Overview}
For our experiments, we utilize the FinQA dataset, which consists of 8,281 pairs of financial questions and answers, crafted by financial experts. The dataset is divided into training (6,251 pairs), validation (883 pairs), and testing (1,147 pairs) sets, in a 75\%/ 10\%/ 15\% split, respectively. Analysis shows that 23.42\% of questions can be answered using text only, 62.43\% require information from tables, and 14.15\% need both text and table data for answers. In terms of reasoning complexity, 59.10\% of the programs involve a single reasoning step, 32.71\% require two steps, and 8.19\% necessitate three or more steps. \newline
In the preliminary section, we've defined 'gold cases' as those with a program score higher than 0.9. When analyzing the top 100 candidates from the dataset based on question similarity, we find that 5.3\% of the training data lacks gold cases for its questions, indicating some questions do not have a matching candidate case with similar questions and programs. 20.6\% of the training set contains fewer than 10 gold cases, while the remaining 74.0\% have more than 10 gold cases. When considering the entire training set (6,250 pairs), only 0.8\% of questions have no gold cases.

\subsection{Experiment on Case Retriever}
The case retriever is crucial for finding the most relevant, or 'gold', cases to support the program generator. We use precision as the key metric to evaluate the case retriever, which assesses how accurately it identifies relevant cases from the dataset in its top-\emph{k} retrieved cases. Precision is the fraction of retrieved gold cases among the top-k retrieved cases, serving as a direct measure of the retriever’s performance.

\subsubsection{Influential Factors on Case Retriever Performance}
The performance of the case retriever is subject to various factors:

\paragraph{Architecture}
We explored two separate architectures: the Bi-encoder and the Cross-encoder, using BERT-base, RoBERTa-base, and RoBERTa-large as pre-trained models for both architectures. Initial findings suggest that the Cross-encoder marginally surpasses the Bi-encoder in performance, with larger models demonstrating improved results. This emphasizes the significance of both model size and architectural design in attaining superior precision in retrieval tasks.

\paragraph{Input Case Variation}
Experiments were conducted using various input types, including questions only, programs only, and a combination of both. The findings reveal that combining questions and programs as inputs significantly improves retrieval performance. This combination approach allows for a more nuanced understanding and representation of cases, leading to more precise retrievals.

\paragraph{Training Set Configuration}
It is crucial to optimize the training set to train our case retriever model, as training on 6,250 question-candidate pairs for each of the 6,251 questions would be excessively time-consuming. We experimented with different strategies for selecting positive and negative training sets, initially focusing on question similarity and later incorporating sampling methods to introduce variability and balance.

\paragraph{Hyperparameter Optimization}
Fine-tuning hyperparameters such as learning rates, optimizers, batch sizes, and the ratio of positive to negative candidates was necessary for enhancing model performance. This process involved iterative testing and adjustments to identify the optimal settings for our specific task.

\subsubsection{Dual-Stage Architecture Exploration}
Further experimentation will focus on the dual-stage architecture integrating both Bi-encoder and Cross-encoder approaches, as discussed in the methodology section. The idea is to refine the number of cases the bi-encoder retrieves, allowing the cross-encoder to focus on a reduced set of candidates during testing. This approach is expected to fine-tune the dual-stage process, leveraging the strengths of both architectures to optimize performance during test time.

\begin{table*}
  \centering
  \begin{tabular}{l|c}
    \hline
    \textbf{Model Architecture} & \textbf{Top-3 Precision} \\
    \hline
    Cross-Encoder (RoBERTa-base, question)                & 77.03 \\
    Cross-Encoder (RoBERTa-base, question\&program)       & 82.79 \\
  \hline
  \end{tabular}
  \caption{Initial Experimental Results on Case Retriever}
  \label{initial_results_retriever}
\end{table*}

\subsection{Experiment on Program Generator}
The main function of the program generator is to create logical steps and corresponding programs that efficiently address financial questions. Through the use of CBR, we anticipate that it will utilize the retrieved cases to enhance the accuracy and pertinence of the answers produced. We aim for the program generator to exhibit improved reasoning skills and to produce precise, logical programs that closely mirror solutions provided by human experts.

Original FinQA research utilizes two main metrics for evaluation: program accuracy and execution accuracy. Program accuracy assesses whether two symbolic programs are mathematically equivalent, focusing on the logical structure of the generated program compared to a gold standard. Execution accuracy measures the correctness of the final answers produced by executing the generated programs. While execution accuracy can sometimes overestimate performance due to coincidentally correct answers, program accuracy might underreport effectiveness because of its inability to recognize multiple valid solutions to the same problem. \newline
To address these issues and specifically gauge the impact of case-based reasoning (CBR) on logical step generation, we introduce Operator accuracy as an additional metric. Operator accuracy examines the precision in generating correct operators within programs, given the case input, providing insight into the CBR's contribution to enhancing logical reasoning in program generation.

\subsubsection{Influential Factors on Program Generator Performance}
Several key factors impact the accuracy of the program generator:

\paragraph{Architecture}
In line with the methodology section, we investigate both Concatenation and Separate Encoder methods for incorporating retrieved cases into the program generation process. The initial findings presented in Table \ref{initial_results_generation} indicate that the Concatenation approach outperforms the Separate Encoder method significantly. This suggests that the model is more effective when using a unified embedding that combines the query's contextual details with supportive case information. To further understand their effectiveness, we will evaluate each strategy using various pre-trained encoder models, including RoBERTa-base and RoBERTa-large. Our goal through this examination is to identify the architecture and model pairing that best facilitates the creation of precise and logically coherent programs.

\paragraph{Input Case Variation}
The choice of input cases plays a complex role in how well the model works. We plan to experiment with different input configurations to understand their impact comprehensively. This includes deciding between using gold cases versus retrieved (noisy) cases in training phase, which might introduce variability. Additionally, we will test whether incorporating both the question and program from cases, or just the program, affects the program generator's effectiveness. The number of cases used as input and the choice between including only operators or the entire program in the input will also be examined. These experiments will shed light on the optimal way to leverage case information for improving program generation.

Through these experimental investigations, we aim to refine our understanding of how best to harness the potential of CBR in enhancing the capabilities of program generators for financial QA tasks. By meticulously examining the effects of architectural choices, encoder models, and input case configurations, we anticipate identifying strategies that significantly improve the generation of logical and accurate programs.

\section{Discussion}
Our research has primarily focused on augmenting the program generator with additional, contextually similar cases to address the challenge of generating incorrect operations, a notable issue in processing complex, real-world documents containing text, tables, and numerical data.

The significant contribution of our study lies in providing the program generator with extra information through similar cases. This methodology is based on the premise that a deeper understanding of multi-modal information can substantially benefit not just the financial domain but also various fields where data is presented in mixed formats. The initial results, particularly the program generator's improved performance with gold cases, underscore the potential of our approach to refine the model's reasoning capabilities and align its outputs more closely with expert-level solutions.

However, our experiments also highlight a critical challenge: enhancing the case retriever's performance. The effectiveness of our proposed method is directly tied to the quality of cases retrieved. As such, refining the case retriever not only stands as our main challenge but also as the primary area for future exploration. The forthcoming stages of this research will aim to quantify the impact of the case retriever's performance on the overall efficacy of the program generator. By focusing on this aspect, we expect to gain valuable insights into optimizing the case retrieval process, thereby maximizing the potential benefits of our CBR approach.

In conclusion, this thesis proposes a novel method of leveraging similar cases to improve the program generator's accuracy in financial QA tasks, with the potential to extend these benefits across various domains dealing with multi-modal data. The dependency of the program generator's success on the case retriever's performance highlights an essential area for further research. As we continue to refine our approach, we aim to not only address the current limitations but also to pave the way for advancements in machine understanding of complex documents, thereby contributing to the broader field of artificial intelligence.

\bibliography{reference}
\bibliographystyle{acl_natbib}

\appendix



\end{document}